\documentclass{article} 
\usepackage{iclr2020_conference,times}
\usepackage[utf8]{inputenc}

\usepackage[arabic,USenglish]{babel}

\usepackage{hyperref}
\usepackage{url}

\title{TUNIZI: a Tunisian Arabizi sentiment analysis Dataset}

\author{Chayma Fourati  \\
iCompass\\
49, Street of Marseille, Tunisia  \\
\texttt{chayma@icompass.digital} \\
\AND
Abir Messaoudi\\
iCompass\\
49, Street of Marseille, Tunisia  \\
\texttt{abir@icompass.digital} \\
\AND
Hatem Haddad \\
iCompass \\
49, Street of Marseille, Tunisia  \\
\texttt{hatem@icompass.digital}
}

\iclrfinalcopy 
\begin{document}

\maketitle

\begin{abstract}
On social media, Arabic people tend to express themselves in their own local dialects. More particularly, Tunisians use the informal way called "Tunisian Arabizi".
Analytical studies seek to explore and recognize online opinions aiming to exploit them for
planning and prediction purposes such as measuring the customer satisfaction and establishing sales
and marketing strategies. 

However, analytical studies based on Deep Learning are data hungry. On the other hand, African languages and dialects are considered low resource languages. For instance, to the best of our knowledge, no  annotated Tunisian Arabizi dataset exists. 
In this paper, we introduce TUNIZI a sentiment analysis Tunisian Arabizi Dataset, collected from social networks, preprocessed for analytical studies and annotated manually by Tunisian native speakers. 

\end{abstract}

\section{Tunisian Arabizi}
Tunisian Arabizi represents the Tunisian arabic text 
written using Latin characters and numerals rather than Arabic letters. 
To illustrate more, some examples of Tunisian Arabizi words with the translation to MSA\footnote{Modern Standard Arabic} and to English are presented in Table \ref{words}.

\begin{table}[!htbp]
\caption{Examples of Tunisian Arabizi common words translated to MSA and English}
\label{words}
\begin{center}
\begin{tabular}{cccc}
\multicolumn{1}{c}{\bf Tunisian Arabizi}  &\multicolumn{1}{c}{\bf MSA} &\multicolumn{1}{c}{\bf English }
\\ \hline \\
3asslema       &  \textRL{\foreignlanguage{arabic}{مرحبا}}   & Hello \\
chna7welek      &   \textRL{\foreignlanguage{arabic}{كيف حالك}}    & How are you \\
sou2el           & \textRL{\foreignlanguage{arabic}{سؤال}}  & Question \\
5dhit           &  \textRL{\foreignlanguage{arabic}{أخذت}} & I took \\
\end{tabular}
\end{center}
\end{table}

In \citet{survey}, a survey was conducted  to address the availability of Tunisian Dialect datasets. The authors concluded that all the existing Tunisian datasets are using Arabic letters and that there is a lack of Tunisian Arabizi annotated datasets.

\citet{parallelcorpus} presented a multidialectal parallel corpus of five Arabic dialects: Egyptian, Tunisian, Jordanian, Palestinian and Syrian in order to identify similarities and possible differences among them. The Overlap Coefficient results, representing the percentage of lexical overlap between the dialects, revealed that the Tunisian dialect has the least overlap with all other Arabic dialects.

These results highlight the problem that the Tunisian Dialect is a low resource language and there is a need to create Tunisian datasets for analytical studies.

In this paper, we present TUNIZI, a Tunisian Arabizi dataset for sentiment analysis studies.  

\section{TUNIZI Dataset}
TUNIZI includes more than 9k Tunisian sentences written only in Tunisian Arabizi. TUNIZI is collected from comments on Youtube social network.
The chosen videos included sports, politics, comedy, TV shows, TV series, arts and Tunisian music videos such that the dataset is representative and contains different types of ages, background, writing, etc.
A script was written by iCompass Team and used in all web scrapping activities.
After collecting the comments, another script was written to remove comments written in French and english. The choice of French and English was made by iCompass team because they represent the most used languages by Tunisians.
However, not all non-Tunisian Arabizi comments were removed so a manual process was conducted in order to make sure the dataset contains only Tunisian Arabizi comments.
\begin{table}[!htbp]
\caption{Tunisian Arabizi comments translated to MSA and English}
\label{comment}
\begin{center}
\begin{tabular}{cccccc}
\multicolumn{1}{c}{\bf Tunisian Arabizi} &\multicolumn{1}{c}{\bf Romanized MSA}&\multicolumn{1}{c}{\bf English }
\\ \hline \\
lkolna m3ak w msendinek & klna m'ek w nsandk & We are all with you and supporting you      \\

nakrhek 5atrek kadheb &  akrhk lank kadb& I hate you because you are a liar
  
\end{tabular}
\end{center}
\end{table}

Table \ref{comment} presents examples of comments with the translation to Romanized MSA and English where the first comment was annotated as positive and the second as negative.

TUNIZI was preprocessed by removing links, emoji symbols and punctuation. 
The annotation task was divided equally on five Tunisian native speakers who annotated the comments as positive or negative.  The three males and two females annotators are at a higher education level (Master/PhD). Two males PhD holders, aged 43 and 42; one female and one male, both aged 25 working as research and development engineers at iCompass and one female aged 23, software engineering student.
An evaluation of the annotation was then performed by the two females in order to make sure annotation was right.

As a result, the dataset is balanced, containing 47\% of positive comments and 53\% negative comments. Thus, statistics, after preprocessing, including the total number of comments, number of positive comments, number of negative comments, number of words and number of unique words are stated in Table \ref{stats}.

\begin{table}[!htbp]
\caption{Dataset Statistics}
\label{stats}
\begin{center}
\begin{tabular}{cc}
\multicolumn{1}{c}{\bf Characteristic}  &\multicolumn{1}{c}{\bf Number } 
 \\ \hline \\
\#Comments  & 9210  \\
\#Negative comments &  4838    \\
 \#Positive comments &  4372     \\
 \#Words&  79862     \\
  \#Unique words&  22850     \\

\end{tabular}
\end{center}
\end{table}

\section{Conclusion}
We presented TUNIZI, the first Tunisian Arabizi Dataset including more than 9K sentences covering different topics, preprocessed and having a balanced sentiment polarity annotation. 
In order to help the African NLP community in further research activities, the dataset is publicly available on the Github link: https://github.com/chaymafourati/TUNIZI-Sentiment-Analysis-Tunisian-Arabizi-Dataset.

As the interest in Natural Language Processing, particularly for African languages, is growing, a natural future step would involve building Arabizi datasets for other underrepresented North African dialects such as Algerian and Moroccan.

\bibliography{africanlpfinal}
\bibliographystyle{iclr2020_conference}

\end{document}